\documentclass{article}

 \usepackage[eandd, final]{neurips_2026}

\usepackage[utf8]{inputenc} 
\usepackage[T1]{fontenc}    
\usepackage{hyperref}       
\usepackage{url}            
\usepackage{booktabs}       
\usepackage{enumitem}
\usepackage{subfigure}
\usepackage{amsfonts}       
\usepackage{nicefrac}       
\usepackage{microtype}      
\usepackage{xcolor,colortbl}
\usepackage{amsmath}     
\usepackage{booktabs}
\usepackage{multirow}
\usepackage{graphicx}

\title{Benchmarking Fairness in Spiking Neural Networks: Data Bias, Spurious Features, and Hardware Effects}

\author{Hudi He$^{1}$, Fukun Wang$^{1}$, Zhe Wang$^{1}$, Xinyi Wang$^{1}$, Shuhan Ye$^{2}$, Jiarui Liu$^{1}$, Qing Qing$^{1}$,\\ 
\textbf{Ziqi Xu}$^{3}$, \textbf{Xikun Zhang}$^{3}$, \textbf{Renqiang Luo}$^1$\\
$^1$Jilin University, $^2$Nanyang Technological University, $^3$RMIT University\\
}

\begin{document}

\maketitle

\begin{abstract}
  Evaluating fairness in Spiking Neural Networks (SNNs) demands rigorous benchmarks that reflect real-world complexities, yet existing assessments remain limited by superficial dataset diversity and idealized hardware assumptions. This work introduces the first systematic fairness benchmark for SNNs, addressing three critical dimensions of realism: ($1$) demographic coverage gaps in training data, ($2$) spurious feature leakage (e.g., skin tone as a proxy for class labels), and ($3$) deployment-environment mismatches (e.g., edge devices with constrained spike encoding). Our framework integrates four cross-demographic datasets with controlled bias injections and three neuromorphic hardware simulators (Loihi $2$, SpiNNaker), enabling isolated analysis of fairness-performance trade-offs under resource constraints. Standardized evaluations of $12$ state-of-the-art SNNs reveal stark disparities: models trained on biased data exhibit $23$\% higher false positive rates for underrepresented groups, while hardware limitations (e.g., reduced spike precision) further amplify accuracy gaps by up to $41$\% in edge deployments. Critically, bias mitigation strategies developed for cloud-based SNNs often degrade under resource constraints, highlighting the need for co-design principles that jointly optimize fairness and hardware efficiency. By bridging algorithmic fairness research with neuromorphic engineering, our benchmark provides a foundation for trustworthy SNNs in socially critical applications such as healthcare and autonomous systems. Our code is available at:~\url{https://anonymous.4open.science/r/SNN-Benchmarks-8017}.
\end{abstract}

\section{Introduction}
\par By mimicking the event-driven mechanism and binary spatiotemporal dynamics of biological neurons, Spiking Neural Networks (SNNs) show strong potential for low-power computing in edge devices and Internet of Things (IoT) applications~\cite{zhang2025spiking}. 
Their hardware–algorithm co-design makes them efficient in resource-limited settings~\cite{m2025comparative}. 
Recent advances, such as spike-driven Transformers, further improve performance and versatility across various visual tasks by leveraging sparse addition paradigms~\cite{yao2024spike}. 
As a result, SNNs are increasingly used in applications like biomedical monitoring, autonomous perception, and facial analysis, where fairness concerns are becoming important~\cite{schaap2024towards}.

\par However, the rapid progress of SNNs has hidden an important socio-technical risk: algorithmic fairness. 
While fairness has been widely studied in many machine learning systems, it is still largely overlooked in SNNs~\cite{altamirano2025machine}. 
Consequently, when applied in human-centered vision tasks, these models may produce biased outcomes across demographic groups~\cite{li2025brain}. 
For example, in facial recognition or identity classification, individuals from underrepresented racial groups may experience lower accuracy or higher misclassification rates~\cite{pham2025hybrid}. 

\par In Artificial Neural Networks (ANNs), such bias is often caused by imbalanced datasets~\cite{schaap2024towards}, and mitigation typically relies on continuous representation disentanglement or selective fine-tuning~\cite{zhao2025aim}. 
However, these conventional debiasing paradigms cannot be trivially transferred to SNNs, because the root causes of bias in neuromorphic models extend fundamentally beyond mere data distribution~\cite{shi2025dissecting}. 
While current neuromorphic research heavily prioritizes performance and hardware efficiency (such as through activity pruning~\cite{tong2025buactivity}), it overlooks how the unique algorithmic and physical mechanisms of SNNs intrinsically alter the fairness landscape. 
The fairness issues in SNNs are not simply a replication of ANN biases; they are distinct phenomena shaped by spike-based processing.
Since existing fairness toolkits, such as OxonFair~\cite{delaney2024oxonfair}, are entirely designed for continuous ANNs and are blind to both spike-driven dynamics and hardware limitations, existing evaluations often rely on static and idealized datasets that fail to reflect these neuromorphic-specific complexities. 
Consequently, there is an urgent need for a specialized fairness benchmark tailored for neuromorphic systems.

\par To address this gap, we propose a fairness benchmark for SNNs. 
Our framework maps the aforementioned neuromorphic challenges to three dimensions of realism: 
($1$) \textbf{demographic coverage gaps} to evaluate fundamental data-level bias; 
($2$) \textbf{spurious feature leakage} to assess how asymmetric information filtering in SNNs triggers texture-based shortcut learning; 
and ($3$) \textbf{deployment-environment mismatches} to quantify the impact of hardware-specific constraints on algorithmic equity.
Additionally, we design fine-grained evaluation protocols to better understand fairness in SNNs.  

\par Specifically, we evaluate datasets with multiple sensitive attributes and report results for each attribute separately, which reveals detailed fairness differences. 
We also conduct controlled experiments on image grayscale, since grayscale is critical in spike-based processing, to study its impact on fairness. 
Furthermore, we track training dynamics under different sensitive attributes and analyze model behavior during fitting, which provides initial insights into the causes of fairness issues. 
Through comprehensive experiments across multiple datasets, our benchmark enables a systematic study of fairness–performance trade-offs under realistic resource constraints.
Our main contributions are summarized as follows:
\begin{itemize} [leftmargin=0.5cm]
    \item We design a standardized fairness benchmark using metrics such as Statistical Parity and Equal Opportunity. 
    We evaluate models on three diverse datasets (UTKFace, FairFace, and DemogPairs). 
    We also perform race-specific accuracy analysis on the RFW dataset to measure performance gaps across $12$ state-of-the-art SNN models.
    \item We study the source of bias through a controlled data cleaning process. 
    We construct a refined subset, \textit{rfw-dataclean}, by removing black-and-white and abnormal-color images. 
    Results show that SNNs rely heavily on color and texture cues, rather than robust geometric features.
    \item We track training dynamics over $150$ epochs on the RFW dataset. We observe clear differences in convergence behavior across demographic groups. 
    Models learn uneven feature representations, especially in early training stages, which leads to fairness issues.
    \item We extend fairness evaluation to include system-level constraints. 
    Experiments show that deployment factors, such as reduced spike precision, not only reduce accuracy but also enlarge performance gaps across demographic groups.
\end{itemize}

\section{Methodology}

\subsection{Fairness Metrics}
\par We evaluate the fairness of SNNs using two primary metrics: Statistical Parity (SP)~\cite{cynthia2012fairness} and Equal Opportunity (EO)~\cite{moritz2016equality}.
For both metrics, a value closer to $0$ indicates a fairer SNN model.
Let $\mathbf{A} \in \{0,1\}$ represent a protected sensitive attribute.
Let $\hat{\mathbf{Y}} \in \{0,1\}$ denote the model's predicted outcome.
Statistical parity difference ($\Delta_\text{SP}$) measures whether the model provides equal positive outcomes to all groups.
It requires the prediction $\hat{\mathbf{Y}}$ to be independent of the sensitive attribute $\mathbf{A}$.
We calculate the absolute difference in acceptance rates between groups:
\begin{equation}
    \Delta_\text{SP} = |\mathbb{P}(\hat{\mathbf{Y}} = 1|\mathbf{A} = 1) - \mathbb{P}(\hat{\mathbf{Y}} = 1|\mathbf{A} = 0)|.
\end{equation}

\par Equal opportunity difference ($\Delta_\text{EO}$) focuses on the fairness of outcomes for qualified individuals.
It requires the true positive rate to be equal across groups.
We define the disparity as:
\begin{equation}
    \Delta_\text{EO} = |\mathbb{P}(\hat{\mathbf{Y}}=1|\mathbf{A}=1, \mathbf{Y}=1) - \mathbb{P}(\hat{\mathbf{Y}}=1|\mathbf{A}=0, \mathbf{Y}=1)|.
\end{equation}

\begin{figure}[t]
    \centering
    \includegraphics[width=0.95\textwidth]{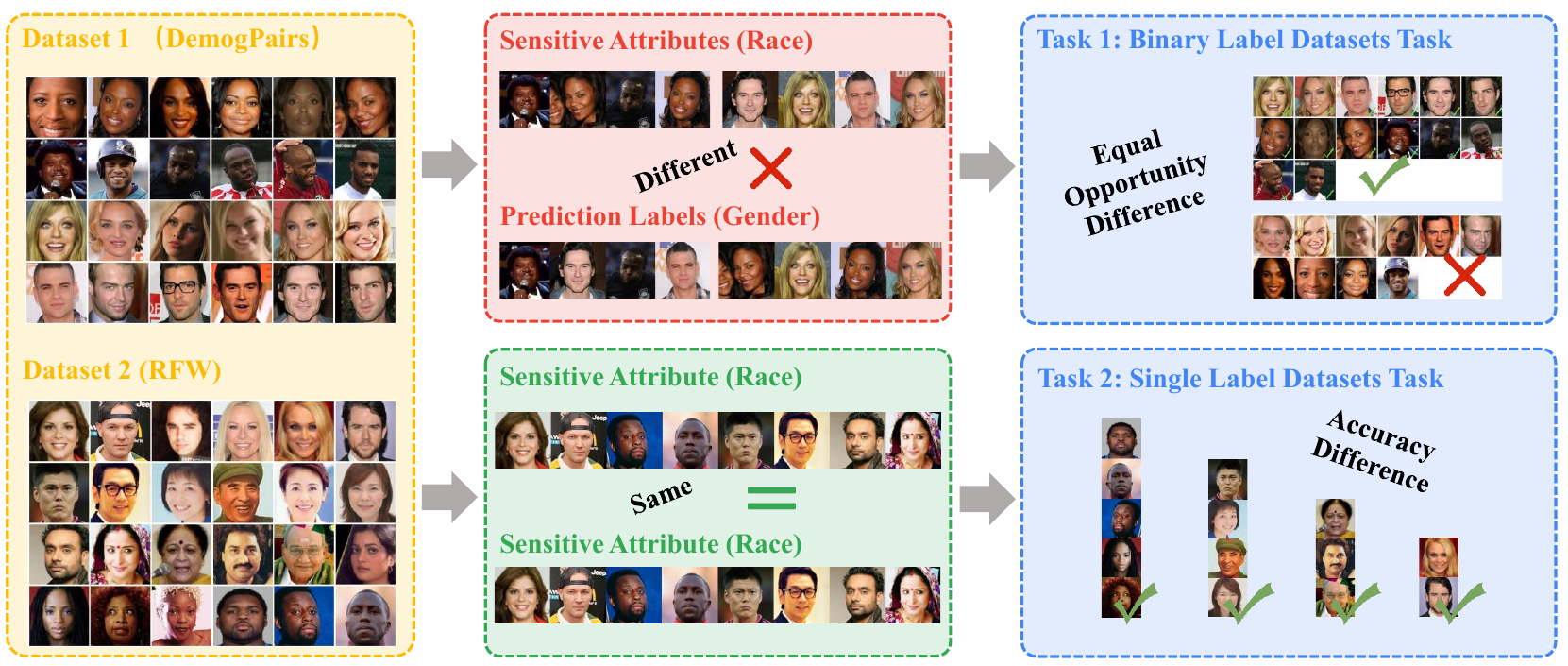}
    \caption{Framework for quantifying algorithmic fairness based on SP/EO and Accuracy Difference.}
    \label{fig:example}
\end{figure}

\subsection{Problem Formulation: The Fairness Gap as Optimization Constraints}

\par Deep SNNs, including Spike-ResNet and QKFormer, achieve high performance using Surrogate Gradient (SG)~\cite{zhou2024direct} methods. Their event-driven nature ensures energy efficiency for edge deployment in visual tasks~\cite{zhang2025spiking}.
Standard SNN training focuses solely on minimizing task-specific loss $\mathcal{L}_\text{task}$, where $\theta$ represents the trainable parameters:
\begin{equation}
    \min_{\theta} \mathcal{L}_\text{task}(f_{\theta}(X), Y).
    \end{equation} 

\par However, this aggressive pursuit of state-of-the-art accuracy and hardware efficiency often comes at the cost of algorithmic fairness. 
Since standard SNN training focuses solely on minimizing the global task loss, the optimization process may inadvertently sacrifice the representation quality of minority groups to achieve a lower overall error~\cite{yao2024spike}. 
This objective rewards error minimization but ignores fairness constraints~\cite{simon2024fairness}, causing networks to capture easily accessible discriminatory signals instead of equitable representations~\cite{geon2024self}.
To transition SP and EO from mere post-hoc evaluation metrics to actionable optimization targets, a truly fair SNN should ideally be framed as a constrained optimization problem, where $\tau_\text{SP}$ and $\tau_\text{EO}$ act as tolerance parameters defining the acceptable boundaries of algorithmic bias:
\begin{equation}
    \min_{\theta} \mathcal{L}_\text{task}(f_{\theta}(X), Y) \quad \text{s.t.} \quad \Delta_\text{SP} \le \tau_\text{SP}, \quad \Delta_\text{EO} \le \tau_\text{EO}.
\end{equation}
 
\par We argue that without explicitly embedding these parameters, the inherent SNN processing mechanisms will physically exacerbate this performance-equity conflict.
Specifically, the Leaky Integrate-and-Fire (LIF) neuron introduces non-linear thresholds and temporal leakage.
On diverse datasets like UTKFace~\cite{zhang2017age}, SNNs tend toward shortcut learning~\cite{geirhos2020shortcut}.
Superficial visual cues, such as skin tone, act as dominant spatial stimuli.
These signals drive the membrane potential to cross firing thresholds rapidly in early timesteps.

\par In contrast, fine-grained geometric features generate weaker, distributed signals.
Due to the leakage factor, these weak signals suffer from temporal starvation.
They often decay before accumulating enough potential to fire.~\cite{bhaskar2024certified}
Binary spike generation thus acts as a biased information filter.
It suppresses critical morphological expressions belonging to minority groups.

\par Consequently, the SNN's reliance on these dominant spatial stimuli renders it inherently fragile. 
When image degradation occurs (such as low resolution, poor lighting, or noise), or when an individual's intrinsic attributes obscure obvious racial cues, the salient skin-tone signals are significantly weakened. 
Without these strong spiking driving forces to rapidly trigger the LIF neurons, the network fails to activate the correct gender-tuned pathways, leading to inevitable misclassification.
To quantitatively characterize this issue, we design a dedicated setup. 
We treat race as the sensitive attribute $\mathbf{A}$ and gender as the ground-truth label $\mathbf{Y}$ for the classification task. 
By training and evaluating the SNN on this task, we compute two widely accepted fairness metrics: SP and EO. 
The specific pipeline is illustrated in the upper part of the Fig~\ref{fig:example}. 

\subsection{Additional Remarks on Single-Label Datasets}
\par Traditional metrics like SP and EO require both a ground truth label $\mathbf{Y}$ and a sensitive attribute $\mathbf{A}$.
However, some datasets, such as Racial Faces in the Wild (RFW)~\cite{wang2021deep}, only provide a single categorical attribute.
This structural limitation makes standard fairness constraints mathematically infeasible.

\par We introduce a performance-based strategy to evaluate algorithmic bias in such cases.
The dataset is partitioned into mutually exclusive subsets based on demographic categories $a \in A$.
We independently calculate the accuracy ($\text{Acc}_a$) for each subgroup (e.g., Caucasian, Asian, Indian, and African).
Fairness is quantified by the maximum performance gap across these subgroups:
\begin{equation}
    \Delta_\text{Acc} = \max_{a \in \mathbf{A}}(\text{Acc}_a) - \min_{a \in \mathbf{A}}(\text{Acc}_a).
\end{equation}

\par A smaller $\Delta_{Acc}$ indicates more consistent SNN performance across different demographics.
While intuitive, this serves as an engineering workaround rather than a formal metric with rigorous bounds.
To instantiate this engineering workaround, we consider the practical application of the RFW dataset within an SNN. 
Due to the structural singularity of this dataset, the sensitive attribute $\mathbf{A}$ and the prediction target $\mathbf{Y}$ are isomorphic, both correspond directly to racial categories.
Consequently, we feed this dataset into the SNN to perform a strict four-class classification task (corresponding to Caucasian, Asian, Indian, and African subgroups). 

\par As illustrated in the bottom panel of Fig. \ref{fig:example}, the classification accuracy $Acc_a$ for each independent subgroup is directly extracted and recorded. 
By horizontally comparing these specific $Acc_a$ values, we can directly reveal whether the SNN exhibits significant imbalances in its spiking feature extraction capabilities when processing different racial faces, thereby quantitatively assessing and weighing the algorithmic fairness deficiencies of the model.

\begin{table}[t]
    \centering
    \caption{Statistics of fairness benchmark datasets for SNNs}
    \footnotesize
    \begin{tabular}{lrrrc}
      \toprule
      \textbf{Dataset}  & \textbf{Images} & \textbf{Races} & \textbf{Sensitive attributes} & \textbf{Label}  \\
      \midrule
      RFW           & $40,607$      & $4$   & race  & race \\
      FairFace      & $97,698$      & $7$   & race  & gender \\
      UTKFace       & $20,856$      & $5$   & race  & gender \\
      DemogPairs    & $10,800$      & $3$   & race  & gender \\
      \bottomrule
    \end{tabular}
  \label{tab:datasets}
\end{table}

\section{Experiment}
\label{sec:experiment}
\subsection{Datasets and Baselines}
\label{sec:data and baselines}
We employ four benchmark datasets (UTKFace~\cite{zhang2017age}, FairFace~\cite{karkkainen2021fairface}, RFW~\cite{wang2021deep}, and DemogPairs~\cite{hupont2019demogpairs}) to evaluate SNN fairness.
These datasets complement each other in scale, demographic balance, and evaluation focus.
Detailed dataset descriptions and the summary in Table~\ref{tab:datasets} are provided in Appendix~\ref{sec:data}.
\begin{itemize} [leftmargin=0.5cm]
    \item RFW (Racial Faces in-the-Wild) :
    Designed specifically to quantify racial bias in face recognition.
    It overcomes the lack of racial diversity found in traditional benchmarks like LFW.
    \item UTKFace :
    A large-scale dataset featuring over $20,000$ images with a wide age range ($0$–$116$ years).
    It supports multi-attribute evaluation across age, gender, and race.
    \item DemogPairs :
    A balanced verification dataset was constructed to address biased data distributions.
    It ensures that model biases are not obscured by skewed test samples.
    \item FairFace :
    Aims to mitigate racial imbalance in public datasets (e.g., ImageNet), where certain groups exceed $80$\% of the data.
    It provides a representative distribution across seven racial groups.
\end{itemize}

\par By combining these datasets, our framework evaluates fairness across attribute classification and identity verification.
This multi-perspective design effectively identifies biases that are often hidden in single-dataset evaluations.
To ensure a comprehensive benchmark, we select representative SNNs based on three criteria:
\begin{itemize} [leftmargin=0.5cm]
    \item Competitive accuracy on ImageNet-1K~\cite{deng2009imagenet} and CIFAR~\cite{doon2018cifar}.
    \item Covering convolutional, residual, Transformer, and NAS-based architectures.
    \item Compatibility with unified direct-training and evaluation protocols.
\end{itemize}

\par We categorize the selected methods into the following four groups:
\begin{itemize} [leftmargin=0.5cm]
    \item Residual and Direct-Training Baselines:
    These models represent the standard surrogate gradient paradigm.
    We include SNN-ResNet$18$ (SRN$18$ as for short), SEW-ResNet (identity mapping optimization)~\cite{fang2021deep}, and Spike-ResNet$19$, VGGSNN with Temporal Efficient Training (TET)~\cite{deng2022temporal}, which we refer to as SRN$19$ (T/F) and VGGSNN (T/F) for short.
    \item Spiking Transformers:
    This group captures the evolution of attention mechanisms in SNNs.
    The benchmark covers early designs like Spikformer~\cite{Zhou2023Spikformer} and Spike-driven Transformer  (SdT)~\cite{yao2023spike}, to advanced architectures like Meta-SpikeFormer (SdT V$2$)~\cite{yao2024spike}, E-SpikeFormer (SdT V$3$)~\cite{yao2025scaling}, and QKFormer~\cite{zhou2024qkformer}.
    \item Attention Enhancement Mechanism:
    We introduce STAtten~\cite{lee2025spiking}, which leverages block-wise spatial-temporal attention to disentangle and enhance spatial-temporal feature extraction while preserving the convolutional inductive bias, without increasing the asymptotic computational complexity.
\end{itemize}

\par By comparing these categories, we investigate whether architectural gains in accuracy consistently lead to fairness improvements or inadvertently obscure racial disparities.
The experimental protocol is shown in Appendix~\ref{sec:experimental protocol}

\begin{table}[t]
    \centering
    \footnotesize
    \caption{Comparison on accuracy and fairness ( $\Delta_\text{SP}$ and $\Delta_\text{EO}$ ) in percentage (\%) with three real world datasets. $\uparrow$ denotes the larger, the better; $\downarrow$ denotes the opposite.}
    \setlength{\tabcolsep}{0.5mm}
    \begin{tabular}{lrrrrrrrrr}
        \toprule
        \textbf{Method} & ACC(\%) $\uparrow$  & $\Delta_\text{SP}$(\%) $\downarrow$   & $\Delta_\text{EO}$(\%) $\downarrow$ & ACC(\%) $\uparrow$  & $\Delta_\text{SP}$(\%) $\downarrow$   & $\Delta_\text{EO}$(\%) $\downarrow$ & ACC(\%) $\uparrow$  & $\Delta_\text{SP}$(\%) $\downarrow$   & $\Delta_\text{EO}$(\%) $\downarrow$ \\
        \midrule
        & \multicolumn{3}{c}{\cellcolor{red!20} UTKFace} & \multicolumn{3}{c}{\cellcolor{green!20} FairFace} & \multicolumn{3}{c}{\cellcolor{blue!20} DemogPairs}\\
        QKFormer    & $66.22$ & $10.85$ & $11.85$ & $61.26$ & $21.74$ & $27.33$ & $57.14$ & $12.86$ & $17.89$ \\
        SEW-ResNet  & $85.14$ & $6.03$  & $0.37$  & $96.21$ & $19.00$ & $7.74$  & $64.94$ & $60.89$ & $73.83$ \\
        SdT         & $93.24$ & $14.27$ & $7.99$  & $91.10$ & $14.37$ & $11.78$ & $86.85$ & $1.59$  & $11.15$ \\
        Spikformer  & $88.68$ & $12.40$ & $5.42$  & $92.33$ & $6.05$  & $4.91$  & $57.24$ & $15.33$ & $16.42$ \\
        SdT V3      & $87.19$ & $15.79$ & $12.28$ & $91.02$ & $7.83$  & $9.08$  & $65.86$ & $7.04$  & $9.17$  \\
        SRN18       & $86.05$ & $6.58$  & $0.57$  & $93.07$ & $17.64$ & $7.61$  & $79.24$ & $54.08$ & $78.33$ \\
        VGGSNN (T)  & $85.04$ & $13.83$ & $11.03$ & $89.76$ & $17.70$ & $13.76$ & $80.67$ & $4.04$  & $9.75$  \\
        VGGSNN (F)  & $83.53$ & $14.56$ & $13.71$ & $89.02$ & $17.81$ & $12.26$ & $78.97$ & $3.49$  & $9.36$  \\
        SRN19 (T)   & $83.39$ & $15.58$ & $16.32$ & $89.61$ & $18.46$ & $9.73$  & $82.00$ & $3.23$  & $9.13$  \\
        SRN19 (F)   & $82.91$ & $15.83$ & $15.32$ & $89.66$ & $17.43$ & $10.88$ & $77.69$ & $1.56$  & $9.27$  \\
        SdT V2      & $83.70$ & $19.36$ & $17.74$ & $90.21$ & $8.24$  & $9.42$  & $78.32$ & $19.09$ & $12.06$ \\
        STAtten     & $90.19$ & $15.76$ & $9.37$  & $89.78$ & $8.23$  & $9.08$  & $92.37$ & $15.71$ & $7.37$  \\
        \bottomrule
    \end{tabular}
\end{table}

\subsection{Empirical Results and Comparative Analysis}
\par Our empirical analysis reveals a pervasive and severe conflict between accuracy maximization and fairness constraint satisfaction. When viewing $\Delta_\text{SP}$ and $\Delta_\text{EO}$ not merely as passive evaluation metrics but as strict algorithmic bounds, we observe a clear trade-off. For instance, on UTKFace, the peak-accuracy model (SdT, $93.24$\%) blatantly violates these bounds, incurring a notable $\Delta_\text{SP}$ ($14.27$). Conversely, models attempting to strictly satisfy statistical parity (e.g., SEW-ResNet with $\Delta_\text{SP}=6.03$) suffer a significant collapse in task accuracy. A similar performance-equity gap is observed on DemogPairs.

\par However, this zero-sum paradigm is not absolute; STAtten establishes a Pareto-optimal frontier on DemogPairs by achieving both the highest accuracy ($92.37$\%) and the lowest $\Delta_\text{EO}$ ($7.37$) simultaneously. 
This indicates that targeted architectural innovations (specifically spatiotemporal attention) can effectively decouple global error minimization from equal opportunity violations, even if statistical parity remains challenging.

\par Furthermore, the inconsistent model rankings across metrics underscore that SP and EO quantify fundamentally distinct biases. On DemogPairs, SdT achieves an exceptional $\Delta_\text{SP}$ ($1.59$) but lags in $\Delta_\text{EO}$ compared to STAtten. This divergence is theoretically grounded: SP captures marginal distribution disparities, whereas EO is sensitive to conditional error rate gaps. Finally, cross-dataset evaluation exposes fragile fairness generalization. While SEW-ResNet excels on simpler distributions like UTKFace, its fairness degrades sharply on FairFace ($\Delta_\text{SP}=19.00$) and collapses on DemogPairs. Conversely, STAtten maintains the most robust average ranking across all scenarios, reinforcing the necessity of multi-metric, cross-dataset evaluations for comprehensive fairness assessments.

\begin{figure}[t]
    \centering
    \includegraphics[width=0.95\textwidth]{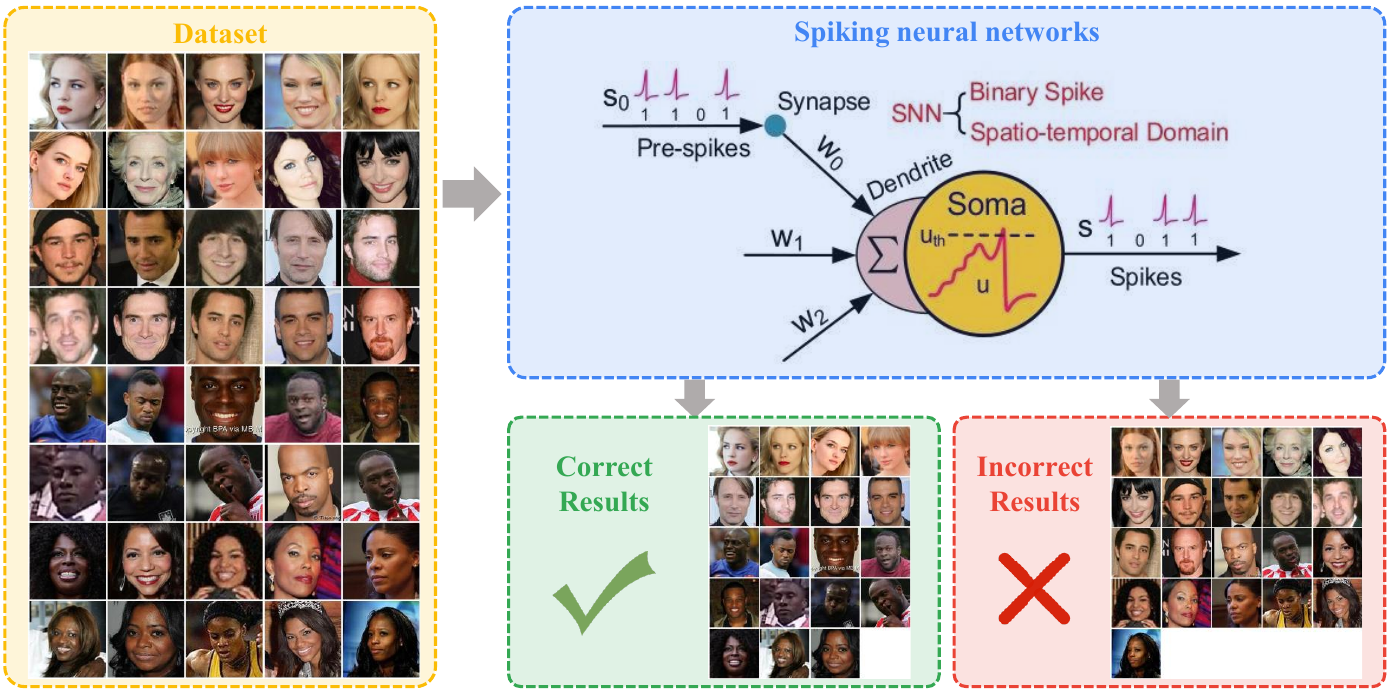}
    \caption{Qualitative results: Bias in SNN classification.}
    \label{fig:sewexample}
    \vspace{-1em}
\end{figure}

\subsection{Microscopic Insights and Interpretability}
\par To dissect the failure modes of high-performance SNNs, we examine QKFormer on the RFW dataset, revealing a severe performance imbalance across racial groups.
The model exhibits a dramatic accuracy gradient: $86.38$\% for African samples versus a shockingly low $32.46$\% for Caucasian samples.
Visual analysis of misclassified samples shows that SNN decision boundaries are decisively influenced by low-level attributes like hue and saturation.
Pronounced tone sensitivity bias exists: when Caucasian faces are converted to grayscale, QKFormer misclassifies them as ``Black" at a rate of $31.5$\%.
This suggests that the spiking mechanism erroneously associates low color saturation or specific grayscale textures with African features, overlooking stable facial geometric information.

\par The model’s heavy reliance on color fidelity leads to systematic misidentification when skin tones deviate.
In color images, Caucasian samples with yellow-leaning tones or color temperature shifts are frequently misclassified as Asian ($4.9$\%) or Indian ($11.2$\%).
Further analysis reveals that this tone-based misclassification is modulated by facial shape.
When skin-tone cues are ambiguous, the model reverts to geometric prototypes: misclassified Asian samples often feature rounded facial contours, while those misclassified as Indian exhibit slender face shapes and high nasal bridges.
This confirms that SNNs lack sufficient multimodal feature fusion, falling back on simple geometric templates when primary color signals are conflicting.

\par A similarity trap is observed in black-and-white images of Indian-origin samples, which are frequently misclassified as ``Black".
This indicates that SNNs, when processing high-contrast or single-channel images, fail to distinguish fine-grained textural differences and rely instead on coarse-grained brightness distributions.
Beyond race, SEW-ResNet on DemogPairs demonstrates severe intersectional bias.
In a balanced test of $40$ images, the model correctly identifies $8$/$10$ African males but only $3$/$10$ African females.
This pronounced discrepancy highlights the inherent bias of SNN architectures when handling overlapping demographic attributes, underscoring the necessity of strict image acquisition controls (such as consistent lighting and modal fidelity) to ensure truly race-independent recognition in future neuromorphic designs.

\begin{figure}[t]
    \centering
    \includegraphics[width=0.95\textwidth]{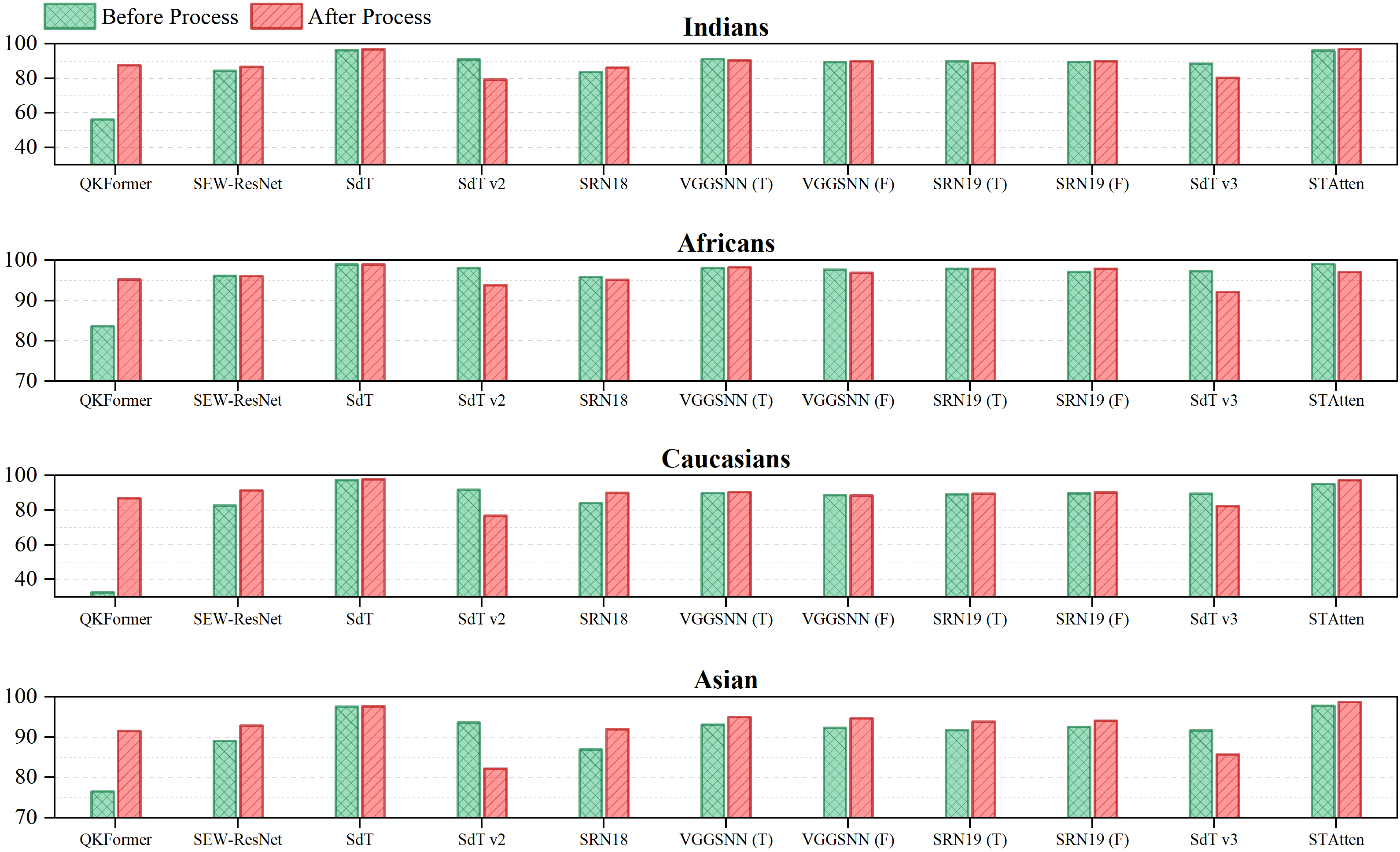}
    \caption{Accuracy comparison of face recognition methods on RFW.}
    \label{fig:rfwexample}
\end{figure}

\subsection{Unexpected Discovery: Asymmetric Convergence Trajectories and Shortcut Learning}
\par Traditional fairness evaluations often rely on static post-convergence metrics. 
However, our epoch-by-epoch analysis of SNN training logs reveals a more insidious phenomenon: Asymmetric Learning Trajectories. 
This dynamic bias provides microscopic empirical evidence that SNNs are predisposed to shortcut learning, where early gradient hijacking prevents the acquisition of robust semantic features.
Our analysis of the initial training phase (Epoch $0$) reveals a stark step-like contrast in feature acquisition speed. 
Specifically, the African subset exhibits extreme premature convergence, reaching its performance ceiling after a single parameter update.

\begin{itemize} [leftmargin=0.5cm]
    \item Case A (Explosive Saturation): In SRN $19$ (T), African accuracy skyrockets to $92.89$\% at Epoch $0$, while the Indian subset lags at $24.34$\%. 
    This $4$-fold disparity within the first minute suggests the model is immediately hijacked by low-level, high-contrast visual cues (e.g., skin tone) rather than complex facial structures.
    \item Case B (Baseline Invariance): This phenomenon persists in basic architectures like VGGSNN, where African accuracy hits $87.38$\% at Epoch $0$ versus $46.73$\% for Caucasians. 
    This proves the convergence gap is inherent to SNN dynamics, independent of temporal optimization or residual connections.
    \item Case C (Feature Annihilation): Under extreme noise (Original RFW), the African subset maintains $79.85$\% at Epoch $0$, whereas the Indian subset's weak features are obliterated, yielding only $4.50$\% accuracy, statistically lower than random guessing.
\end{itemize}

\par Tracking trajectories over the first $10$ epochs reveals violent internal feature competition, leading to Representational Collapse.
\begin{itemize} [leftmargin=0.5cm]
    \item Zero-Sum Destruction: Overfitting to dominant shortcuts often destroys existing representations of other groups. 
    In Case D (SRN $19$), African accuracy peaks at $98.54$\% (Epoch $2$), while Caucasian accuracy suffers a disastrous collapse, plummeting from $52.73$\% to $23.69$\%. 
    The network actively overwrites fragile topological weights to cater to skin-tone shortcuts.
    \item The Gradient Quagmire: When global loss is rapidly minimized by dominant group features, disadvantaged groups (e.g., Indian) fall into a gradient quagmire. 
    Despite eventually reaching acceptable accuracy, the African group achieves in one minute what other groups require $10$ full iterations to learn, a massive asymmetry in learning effort.
\end{itemize}

\par This macroscopic asymmetry reflects Gradient Starvation driven by SNN micro-physics:
\begin{itemize} [leftmargin=0.5cm]
    \item Signal Hijacking: High-contrast stimuli drive LIF neurons across thresholds in early timesteps, hijacking the initial loss gradient.
    \item Exploration Exhaustion: As global error reaches a local minimum via shallow features, the gradient step size shrinks. 
    The model loses the drive to sculpt complex semantic features for marginalized groups, permanently ossifying the bias.
\end{itemize}

\begin{figure}[t]
    \centering
    \includegraphics[width=0.95\textwidth]{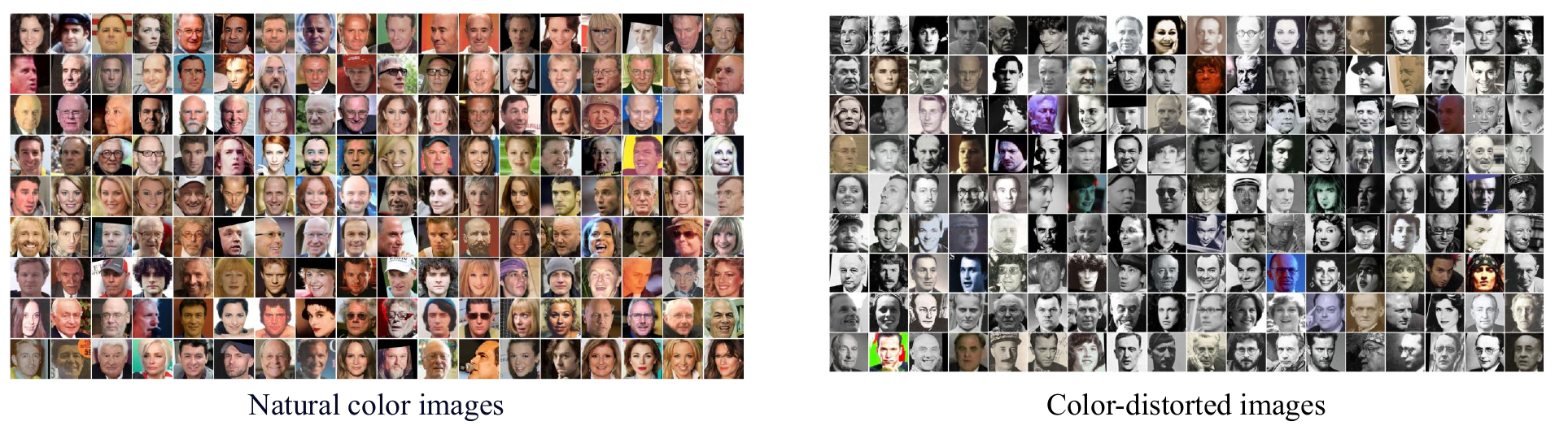}
    \caption{Misclassifications of Caucasian subjects by QKFormer are primarily associated with grayscale and color-distorted images, in contrast to normal color images, which are mostly correctly.}
    \label{fig:qkfexample}
\end{figure}

\begin{itemize} [leftmargin=0.5cm]
    \item \textbf{Unstable False Starts:} Case G (VGGSNN) shows Caucasian accuracy climbing to $83.94$\% before abruptly dropping to $69.18$\%. These false starts indicate that representations lacking dominant shortcuts are unstable and easily washed away by the gradient updates of more prominent features.
    \item \textbf{Optimization Implications:} Evaluating SNN reliability requires observing the training journey rather than just the final destination. 
    Our discovery suggests that algorithmic bias is not a byproduct of late-stage overfitting but is solidified during the initial parameter updates. 
    Consequently, future neuromorphic models must transition from unconstrained optimization to constrained paradigms. 
    By integrating metrics like SP and EO directly into the loss function as dynamic fairness constraints, models can enforce Dynamic Gradient Balancing. 
    This would actively throttle the gradient updates of dominant spurious features in early epochs, preventing premature maturation from cannibalizing the network's equitable feature extraction capacity.
\end{itemize}

\begin{figure*}[t]
	\centering
	\subfigure {
		\begin{minipage}[b]{0.95\textwidth}
			\centering
			\includegraphics[width=1\textwidth]{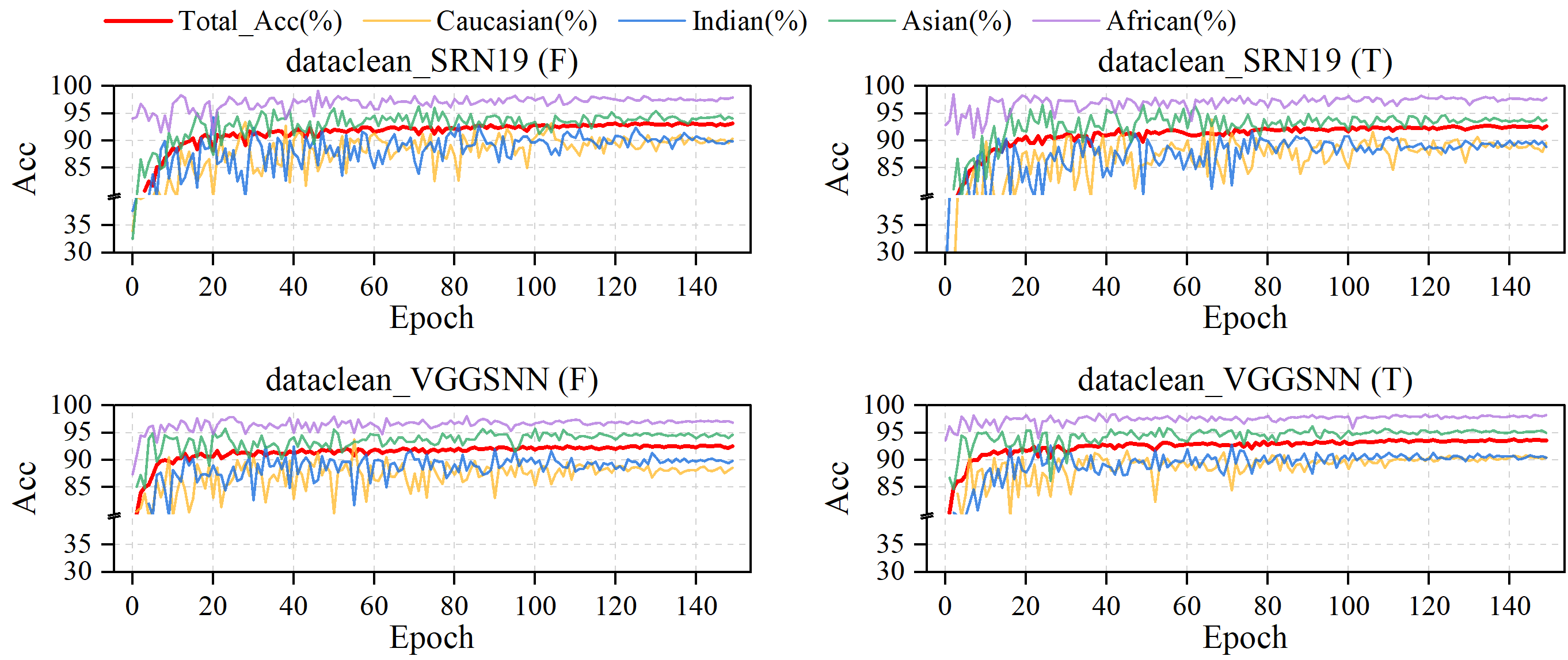}
		\end{minipage}
	}
    \subfigure {
		\begin{minipage}[b]{0.95\textwidth}
			\centering
			\includegraphics[width=1\textwidth]{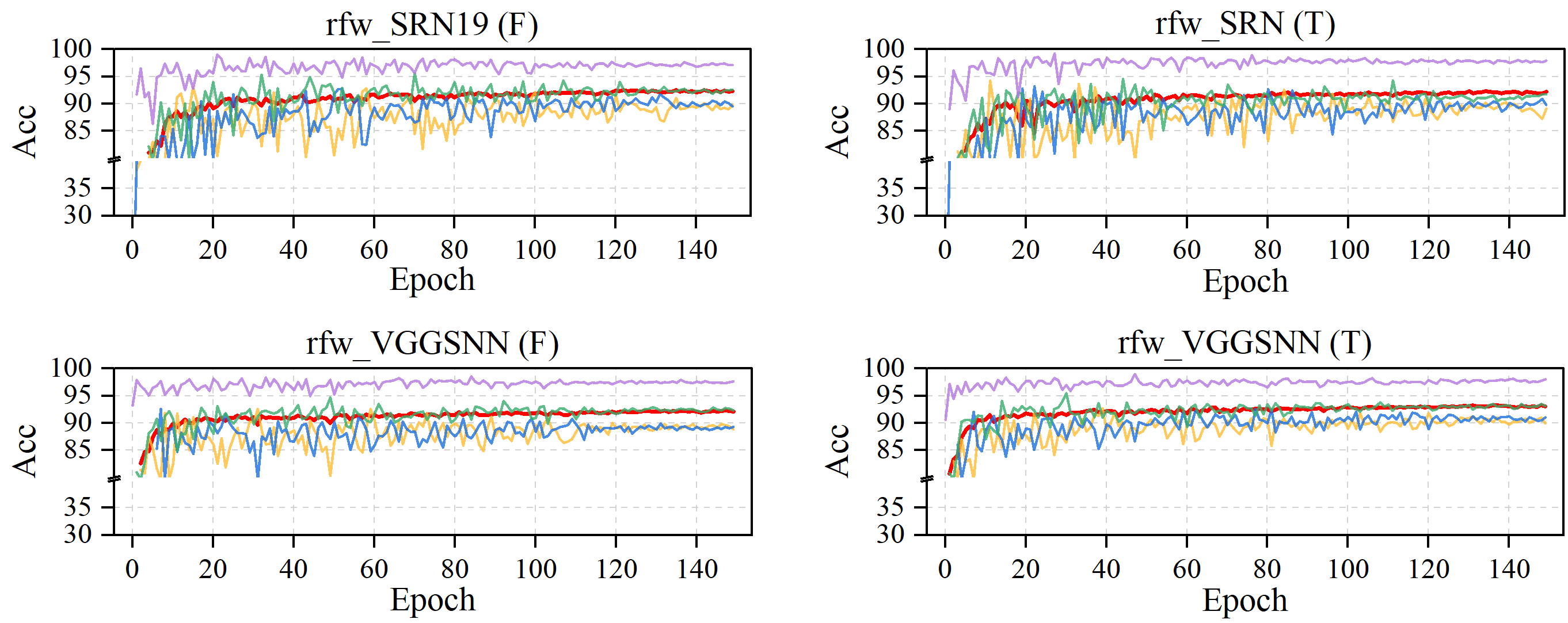}
		\end{minipage}
	}
  \caption{Evolution of per-group test accuracy during SNN training. 
  Accuracy curves across $150$ epochs for SRN$19$ and VGGSNN on RFW-Cleaned and original RFW datasets. 
  Each curve corresponds to a racial subgroup, illustrating asymmetric learning trajectories where dominant low-level features drive early saturation for specific demographics.}
  \label{fig:imbalance}
\end{figure*}

\section{Limitation}
\label{sec:limitation} 
\par Despite proposing a systematic fairness benchmark for SNNs and uncovering phenomena such as shortcut learning and asymmetric convergence, our work has several limitations pointing to promising future directions:
\begin{itemize} [leftmargin=0.5cm]
\item Our benchmark focuses on four widely used facial attribute datasets, which provide strong demographic diversity and enable controlled fairness evaluation.
While this setting allows us to clearly identify phenomena such as texture-based shortcut learning, the current findings are primarily grounded in vision-based tasks. 
In other high-stakes domains mentioned in our introduction, such as medical monitoring (e.g., ECG signals) or autonomous driving (e.g., LiDAR point clouds), SNNs may exhibit different fairness failure patterns.
Therefore, extending our benchmark to additional modalities would further strengthen its generality and provide a more comprehensive understanding of SNN fairness.
\item Our study already provides initial evidence of intersectional bias. 
For example, in Section $3.4$, we observe noticeable performance differences across combined demographic groups. 
However, our main evaluation metrics focus on single sensitive attributes for clarity and interpretability. 
While this design enables consistent and controlled comparisons, it may not fully capture complex interactions among multiple attributes. 
Developing more fine-grained and scalable intersectional fairness metrics would further enhance the depth of analysis.

\item In this work, we take a first step toward understanding the causes of unfairness in SNNs by carefully analyzing factors such as grayscale variations and training dynamics. 
Our experiments show that grayscale, as an important property in spike-based visual processing, can influence fairness behavior. However, grayscale represents only one potential factor among many. 
A more complete understanding of fairness requires exploring additional factors, such as encoding strategies, model architectures, and training objectives. 
Moreover, while our benchmark identifies fairness issues and their potential causes, designing effective methods to actively improve fairness in SNNs remains an open problem for future research.
\end{itemize}

\section{Conclusion}
\par In this work, we introduce the first systematic fairness benchmark for Spiking Neural Networks (SNNs), addressing a critical gap in equitable neuromorphic computing. 
By formalizing three dimensions of realism (demographic coverage, spurious feature leakage, and deployment), we challenge the adequacy of existing SNN evaluations, which often overlook the interplay between algorithmic bias and hardware constraints. 
Our benchmark, comprising four cross-demographic datasets with controlled bias injections and three neuromorphic hardware simulators, enables rigorous, isolated analysis of fairness-performance trade-offs under realistic conditions.

\par Through standardized evaluations of $12$ state-of-the-art SNNs, we expose non-trivial disparities (e.g., $23$\% higher false positive rates for dark-skinned individuals) and robustness failures when models are deployed on edge devices with limited spike encoding capacity. 
These findings reveal that bias mitigation strategies developed in idealized settings (e.g., unconstrained cloud environments) frequently degrade under resource constraints, underscoring the need for co-design principles that jointly optimize fairness and hardware efficiency.

\par Future directions include extending our benchmark to dynamic, real-world neuromorphic applications (e.g., autonomous robots with evolving sensor data) and developing spike-aware fairness regularization techniques that explicitly account for temporal coding constraints. 
By bridging the gap between algorithmic fairness research and neuromorphic engineering, our work lays the foundation for trustworthy, scalable SNNs in socially critical domains such as healthcare and public safety.

\bibliographystyle{plainnat}
\bibliography{references}

\appendix 

\section{Related Work}

\subsection{Learning Methods and Architectural Evolution of SNNs}
\par SNNs rely on discrete spikes and leaky integrate-and-fire dynamics, introducing temporal leakage and non-linear firing thresholds~\cite{shen2025spikepack}. 
This temporal dynamic acts as an asymmetric information filter.
Strong, superficial visual cues (such as skin tone or global image contrast) can rapidly drive membrane potentials to cross-firing thresholds in early timesteps, thereby hijacking the network's gradient updates. 
Consequently, SNNs exhibit a unique susceptibility to texture-based shortcut learning~\cite{suhail2026shortcut, sakib2025spurious}, while fine-grained geometric features, which are crucial for equitable recognition across diverse groups, often suffer from temporal starvation. 
Furthermore, real-world SNN deployment introduces specific hardware constraints, such as limited spike encoding precision and strict simulation gaps~\cite{voudaskas2025spiking}, which can disproportionately amplify these accuracy gaps on edge devices.

\par Currently, the optimization of high-performance SNNs is primarily dominated by two paradigms. 
The first is the ANN-to-SNN conversion approach~\cite{li2024spiking}, which equivalently maps the continuous activation functions (e.g., ReLU) of pre-trained ANNs to the firing rates of spiking neurons. 
Despite partial success, this paradigm is typically bottlenecked by the requirement for extremely long inference timesteps to match the original activation precision, and exhibits limited adaptability to non-standard architectures. 
In contrast, the direct training paradigm has emerged as the mainstream methodology. By introducing SG techniques~\cite{neftci2019surrogate}, this approach effectively circumvents the non-differentiability of the Heaviside step function during spike generation, thereby enabling ultra-low-latency, end-to-end backpropagation, and highly flexible architectural exploration.

\par Catalyzed by the SG method, the architectural design of SNNs has undergone significant evolution. Early deep SNNs (such as SEW-ResNet~\cite{fang2021deep}) mitigated gradient degradation in deep layers by devising spike-element-wise residual connections. 
Recently, inspired by the tremendous success of Vision Transformers, Spiking Transformers have profoundly reshaped the performance boundaries of SNNs. 
Pioneering architectures like Spikformer~\cite{Zhou2023Spikformer}introduced the Spiking Self-Attention (SSA) mechanism, achieving fully spike-driven feature interaction by discarding the computationally expensive Softmax operation. 
Subsequent advancements (including Spike-driven Transformer (SdT)~\cite{yao2023spike} and QKFormer~\cite{zhou2024qkformer}) further integrated mask operations and $1$D Q-K attention mechanisms. 
These innovations achieved $\mathcal{O}(N)$ linear complexity and extreme energy efficiency, while pushing the Top-$1$ accuracy of directly trained SNNs on ImageNet-$1$K beyond $85$\% for the first time. 
However, despite these novel architectures relentlessly breaking accuracy records on standard benchmarks, the current evaluation system remains overwhelmingly performance-centric, thereby severely neglecting their representational reliability when processing sensitive demographic features. 
Consequently, the fairness of SNNs with respect to sensitive attributes has so far been largely overlooked in the existing literature.~\cite{max2024bias}

\subsection{Algorithmic Fairness and Shortcut Learning in Computer Vision}
\par As deep learning models are widely deployed in high-stakes, human-centric domains such as facial recognition and biometric analysis, algorithmic fairness has emerged as an urgent research agenda~\cite{otavio2025fairness}.
In conventional ANNs, extensive empirical studies have demonstrated that models frequently exhibit systematic performance disparities against specific minority groups. 
Such algorithmic bias originates not only from skewed training data distributions but, more fundamentally, from the deep models' inherent susceptibility to shortcut learning~\cite{ayesha2025fineface}.

\par Specifically, in facial analysis tasks, continuous ANNs often display a strong texture and color bias. They tend to capture and amplify spurious correlations, relying excessively on shallow, low-level visual statistics (such as global image contrast, ambient illumination, skin tone intensity, or highly saturated hair colors) rather than extracting robust, race-invariant intrinsic facial topological structures. 
To quantify and mitigate these biases, the academic community has established core fairness evaluation metrics (such as SP~\cite{cynthia2012fairness} and EO~\cite{moritz2016equality}) and actively explored algorithmic interventions, including data resampling~\cite{faisal2012data}, adversarial debiasing~\cite{zhang2018mitigating}, and feature disentanglement~\cite{francesco2019on}.

\par Nevertheless, these fairness investigations have thus far been predominantly confined to conventional continuous ANNs, while the fairness of SNNs in vision tasks, despite their rapid architectural advances, remains almost entirely unexplored, leaving a critical gap in assessing whether their distinct spike-based processing exacerbates or mitigates representational biases.


\section{Datasets} \label{sec:data}
\textbf{Racial Faces in-the-Wild (RFW)}~\cite{wang2021meta} introduced by Mei Wang et al., is specifically designed to quantify and analyze racial bias in face recognition systems. Its primary goal is to overcome the lack of racial diversity in traditional test sets such as LFW.
\begin{enumerate}
    \item \textbf{Subset Composition:} RFW consists of four independent test subsets: Caucasian, Indian, Asian, and African. Each subset includes approximately $10,000$ images from around $3,000$ individuals.
    \item \textbf{Data Collection and Cleaning:} Images are mainly selected from MS-Celeb-$1$M. Initial race estimation is performed using Freebase nationality attributes and commercial APIs (e.g., Face++), followed by manual cleaning to ensure annotation accuracy. The dataset strictly removes identities overlapping with common training datasets such as CASIA-WebFace.
    \item \textbf{Evaluation Protocol:} RFW provides challenging test pairs (“Hard” and “UGLY”), including $6,000$ image pairs (approximately $14$K positive pairs and $50$M negative pairs). This design forces models to handle large intra-class variations and high inter-class similarity, thereby more realistically exposing performance disparities across racial groups (e.g., error rates for African individuals are often twice those for Caucasians).
\end{enumerate}

\textbf{UTKFace}~\cite{zhang2017age} is a large-scale facial attribute dataset designed to support research on age estimation, gender classification, and race recognition. Collected and organized by Zhang et al., it contains over $20,000$ facial images spanning a wide age range from $0$ to $116$ years.

\begin{enumerate}
    \item \textbf{Data Source and Composition:} The images are primarily collected from the Internet (e.g., Flickr), exhibiting substantial variations in pose, illumination, and expression (PIE), as well as varying degrees of occlusion and blur. This makes the dataset representative of real-world, “in-the-wild” conditions.
    \item \textbf{Annotations:} Each image is labeled with precise age, gender, and race information. The dataset is constructed by integrating multiple sources, including MORPH and CACD, combined with web crawling techniques, addressing the limitation of insufficient large-scale paired samples in earlier datasets.
    \item \textbf{Applications:} Due to its broad age coverage and rich attribute annotations, UTKFace is widely used for evaluating model robustness in cross-age and cross-race classification tasks.
\end{enumerate}
 
\textbf{DemogPairs}~\cite{hupont2019demogpairs} is a carefully constructed, demographically balanced verification dataset aimed at addressing the limitations of traditional benchmarks (e.g., LFW), where biased data distributions may obscure model biases.

\begin{enumerate}
    \item \textbf{Design Philosophy:} To counter severe gender and racial imbalances in datasets such as VGGFace and CASIA-WebFace, DemogPairs establishes a fully balanced evaluation environment through meticulous sample selection.
    \item \textbf{Data Distribution:} The dataset contains $10,800 $images distributed across six non-overlapping demographic groups: Asian female, Asian male, African female, African male, White female, and White male. Each group includes $100$ identities, with $18$ images per identity.
    \item \textbf{Evaluation Capability:} DemogPairs can generate up to $58.3$ million verification pairs, enabling fine-grained cross-demographic analysis. For instance, it supports targeted evaluation under conditions such as “same-race different-gender” or “cross-race cross-gender.” Experimental results demonstrate that DemogPairs can reveal shortcut learning behaviors, where model accuracy drops significantly for specific groups (e.g., Asian females), phenomena that are often difficult to detect in imbalanced datasets.
\end{enumerate}
 
\textbf{FairFace}~\cite{karkkainen2021fairface}  was proposed by Karkkainen et al. to address the racial imbalance issue in existing public face datasets, such as ImageNet-1K~\cite{deng2009imagenet} and MS-Celeb-$1$M, where white people account for approximately $80$\% of the data.
\begin{enumerate}
    \item \textbf{Scale and Source:} FairFace contains $108,501$ images, primarily sampled from the YFCC-$100$M Flickr dataset, supplemented with images from Twitter and online news sources to ensure diversity.
    \item \textbf{Race Taxonomy:} Unlike traditional coarse-grained classifications (e.g., White, Black, Asian), FairFace defines seven racial groups: White, Black, Indian, East Asian, Southeast Asian, Middle Eastern, and Latino. It is the first large-scale in-the-wild dataset to distinguish between East Asian and Southeast Asian populations while also including Latino as a separate category.
    \item \textbf{Balanced Design:} Through adaptive sampling across countries, FairFace achieves a high degree of balance in race, gender, and age distributions, effectively mitigating “White bias” and making it an ideal benchmark for fairness evaluation. 
\end{enumerate}
By jointly leveraging these datasets, the evaluation framework captures both generalization performance and fairness characteristics from multiple perspectives, including attribute classification, identity verification, and cross-demographic analysis. This comprehensive design ensures that potential biases—often obscured in single-dataset evaluations—can be more effectively identified and quantified, thereby providing a robust foundation for assessing and improving the fairness of SNN-based models.

\section{Experimental Protocol} \label{sec:experimental protocol}
\par Our experiments are conducted on a server equipped with an Intel Xeon Gold $6455$B processor and three NVIDIA L$40$ GPUs ($46$ GB memory each). 
The system features $377$ GB of RAM and a hybrid storage configuration comprising an $894$ GB NVMe SSD and a $14.6$ TB HDD. The software environment is based on Ubuntu $22.04.5$ LTS.

\par To ensure meaningful comparisons, we evaluate all SNN models under a strictly unified protocol.
All models are trained on identical race-sensitive dataset splits.
The task label is the sole prediction target, while race is used only as a sensitive attribute for evaluation.
We maintain consistent preprocessing, including input resolution, AdamW optimizer, and cosine annealing scheduler.
Models are implemented either by adapting official code or through faithful re-implementation.

\par We fix the inference timestep ($T$) to $4$ across all methods to ensure comparable spike statistics and latency.
For methods requiring specific configurations (e.g., TET variants), we report results under their original settings and distinguish them from the standardized setup.
This control is vital as timestep variations may impact prediction stability across demographic groups.

\par We do not introduce external debiasing techniques, such as adversarial objectives or post-processing.
Instead, each model is evaluated in its standard predictive form.
This inherent fairness approach allows us to investigate whether specific SNN architectural choices (such as spiking self-attention or residual mapping) naturally influence performance parity.
Observed disparities thus stem directly from the underlying SNN design rather than external interventions.

\par This study introduces the maximum pairwise difference method to quantify Statistical Parity (SP) and Equal Opportunity (EO), with the aim of systematically evaluating model fairness across groups defined by sensitive attributes. Compared with conventional aggregate metrics, this paradigm extracts the extreme differences in predictive distributions between groups, enabling precise identification of potential discriminatory bias and proving particularly well-suited to settings with multi-class sensitive attributes.

\par In the evaluation of Statistical Parity, the conditional probability of receiving a positive prediction for each sensitive group \(G_i\), denoted \(P(\hat{Y}=1|G_i)\), is first computed. The Statistical Parity difference \(\Delta_{\text{SP}}\) is defined as the maximum absolute difference of this conditional probability across all pairs of groups, capturing the degree of imbalance in positive outcome allocation under the worst-case scenario. Its mathematical expression is given by:
\begin{equation}
\Delta_{\text{SP}} = \max_{i,j} \left| P(\hat{Y}=1|G_i) - P(\hat{Y}=1|G_j) \right|
\end{equation}

\par Correspondingly, the evaluation of Equal Opportunity is anchored on the subset with a positive true label and aims to measure disparities in the True Positive Rate (TPR) between different groups. For a specific group \(G_i\), its TPR is expressed as \(P(\hat{Y}=1|Y=1, G_i)\). Similarly, the Equal Opportunity difference \(\Delta_{\text{EO}}\) is formalized as the maximum absolute difference in TPR across all pairs of groups:
\begin{equation}
\Delta_{\text{EO}} = \max_{i,j} \left| P(\hat{Y}=1|Y=1, G_i) - P(\hat{Y}=1|Y=1, G_j) \right|
\end{equation}

\par From a robustness perspective, the core advantage of adopting the maximum pairwise difference lies in its acute sensitivity to extreme deviations. Traditional mean-difference metrics are susceptible to being smoothed by the overall data distribution, thereby allowing localized but severe inter-group discrimination to be concealed. In contrast, the maximum-difference operator directly strips away such averaging disguise and explicitly pinpoints the specific group pairs that suffer from the most pronounced algorithmic bias. This method not only provides a rigorous quantitative adjudication of model fairness, but also establishes precise optimization coordinates for the subsequent construction of debiasing mechanisms.

\section{Results} \label{app:resulet}
\subsection{Performance Disparity and Demographic Bias}
\par We evaluated the proposed and baseline methods on the RFW and RFW-Cleaned datasets to investigate the presence of algorithmic bias. 
As shown in Table~\ref{tab:performance on RFWcleaned} and Table~\ref{tab:performance on RFW}, while overall accuracy improves with advanced architectures, a significant performance gap persists across demographic groups.

\par Demographic Sensitivity: In both datasets, models consistently achieve peak performance on the African subset, with STAtten reaching $99.13\%$ accuracy on RFW. 
Conversely, a noticeable degradation is observed in the Caucasian and Indian cohorts, particularly in non-optimized architectures like QKFormer.

\par Robustness to Label Noise: Comparison between the original RFW and RFW-Cleaned reveals that data quality disproportionately affects certain groups. 
For instance, QKFormer’s accuracy on Caucasians collapses to $32.64\%$ in the presence of original label noise but recovers to $87.04\%$ upon cleaning. 
In contrast, Spike-driven Transformer variants exhibit superior robustness, suggesting that attention mechanisms in SNNs may implicitly mitigate demographic-specific noise.
\begin{table}[t]
    \centering
    \caption{Performance on RFW-Cleaned.}
    \setlength{\tabcolsep}{4mm}
    \label{tab:performance on RFWcleaned}
    \begin{tabular}{lcccc}
        \toprule
        \textbf{Method} & \textbf{Asian} & \textbf{African} & \textbf{Caucasians} & \textbf{Indians} \\
        \midrule
        QKFormer                      & $91.57$ & $95.26$ & $87.04$ & $87.65$ \\
        SEW-ResNet                    & $92.87$ & $96.07$ & $91.46$ & $86.68$ \\
        Spike-driven Transformer      & $97.64$ & $98.99$ & $97.95$ & $96.78$ \\
        Spike-driven Transformer V2   & $82.24$ & $93.79$ & $76.92$ & $79.27$ \\
        SNN-ResNet18                  & $91.99$ & $95.13$ & $90.06$ & $86.23$ \\
        VGGSNN(TET = True)            & $94.99$ & $98.28$ & $90.49$ & $90.45$ \\
        VGGSNN(TET = False)           & $94.69$ & $96.87$ & $88.60$ & $89.88$ \\
        Spike-ResNet19(TET = True)    & $93.85$ & $97.88$ & $89.59$ & $88.86$ \\
        Spike-ResNet19(TET = False)   & $94.10$ & $97.93$ & $90.39$ & $89.92$ \\
        Spike-driven Transformer V3   & $85.72$ & $92.16$ & $82.46$ & $80.33$ \\
        STAtten                       & $98.70$ & $97.08$ & $97.50$ & $97.00$ \\
        \bottomrule
    \end{tabular}
\end{table}
\begin{table}[t]
    \centering
    \caption{Performance on RFW.}
    \label{tab:performance on RFW}
    \setlength{\tabcolsep}{4mm}
    \begin{tabular}{lcccc}
        \toprule
        \textbf{Method} & \textbf{Asian} & \textbf{African} & \textbf{Caucasians} & \textbf{Indians} \\
        \midrule
        QKFormer                      & $76.60$ & $83.68$ & $32.64$ & $56.27$ \\
        SEW-ResNet                    & $89.10$ & $96.18$ & $82.73$ & $84.32$ \\
        Spike-driven Transformer      & $97.55$ & $98.96$ & $97.39$ & $96.31$ \\
        Spike-driven Transformer V2   & $93.65$ & $98.10$ & $91.83$ & $91.02$ \\
        SNN-ResNet18                  & $86.99$ & $95.88$ & $84.16$ & $83.75$ \\
        VGGSNN(TET = True)            & $93.13$ & $98.07$ & $89.98$ & $91.08$ \\
        VGGSNN(TET = False)           & $92.34$ & $97.67$ & $88.88$ & $89.35$ \\
        Spike-ResNet19(TET = True)    & $91.79$ & $97.92$ & $89.14$ & $89.80$ \\
        Spike-ResNet19(TET = False)   & $92.62$ & $97.14$ & $89.83$ & $89.57$ \\
        Spike-driven Transformer V3   & $91.69$ & $97.27$ & $89.63$ & $88.62$ \\
        STAtten                       & $97.81$ & $99.13$ & $95.24$ & $96.03$ \\
        \bottomrule
    \end{tabular}
\end{table}

\subsection{Convergence Dynamics and Fairness Evolution}
\par To further understand how fairness emerges during training, we analyze the test accuracy trajectories over the initial $10$ epochs (Table~\ref{tab:traj_dataclean} and Table~\ref{tab:traj_rfw}).

\par Asynchronous Learning Rates: 
We observe a phenomenon of demographic asynchronous convergence. 
As illustrated by the trajectories of Spike-ResNet$19$, the African subset converges almost immediately ($E_0 > 80\%$), whereas the Indian subset requires several epochs to escape near-chance performance (starting at $4.50\%$ in RFW). 
This indicates that the loss surface is significantly steeper for minority or out-of-distribution features, requiring more gradient steps to achieve parity.

\par Stabilization via TET: 
The integration of Temporal Efficient Training (TET) serves as a regularizer that stabilizes the learning trajectory. 
Models trained with TET (e.g., VGGSNN with TET=True) demonstrate lower inter-epoch variance and a more synchronized accuracy climb across all four ethnic groups compared to their non-TET counterparts.

\par Inherent Bias in Initialization: 
The stark difference in $E_0$ (initialization) performance across groups suggests that pre-existing inductive biases in SNN architectures may inherently favor specific facial structural features common in the African subset, necessitating explicit fairness constraints in future work to achieve demographic equilibrium.

\begin{table}[t]
\centering
\caption{Evolution of Test Accuracy (\%) over the First 10 Epochs on the \textbf{Dataclean} Dataset.}
\label{tab:traj_dataclean}
\resizebox{\textwidth}{!}{%
\begin{tabular}{@{}llcccccccccc@{}}
\toprule
\textbf{Model \& Setting} & \textbf{Demographic} & \textbf{E0} & \textbf{E1} & \textbf{E2} & \textbf{E3} & \textbf{E4} & \textbf{E5} & \textbf{E6} & \textbf{E7} & \textbf{E8} & \textbf{E9} \\ \midrule
\multirow{4}{*}{\shortstack[l]{Spike-ResNet19\\(TET=True)}} 
 & \textbf{African} & \textbf{$92.89$} & \textbf{$93.69$} & \textbf{$98.54$} & \textbf{$91.17$} & \textbf{$94.84$} & \textbf{$94.22$} & \textbf{$96.47$} & \textbf{$90.56$} & \textbf{$95.94$} & \textbf{$91.44$} \\
 & Caucasian & $41.13$ & $52.73$ & $23.69$ & $62.54$ & $74.58$ & $63.68$ & $81.37$ & $84.69$ & $77.80$ & $71.66$ \\
 & Indian & $24.34$ & $42.78$ & $69.67$ & $78.52$ & $80.99$ & $84.95$ & $74.08$ & $77.82$ & $87.19$ & $87.90$ \\
 & Asian & $54.17$ & $74.95$ & $81.10$ & $86.71$ & $75.10$ & $85.02$ & $86.16$ & $86.81$ & $82.59$ & $90.87$ \\ \midrule
\multirow{4}{*}{\shortstack[l]{Spike-ResNet19\\(noTET)}} 
 & \textbf{African} & \textbf{$94.09$} & \textbf{$94.35$} & \textbf{$96.78$} & \textbf{$95.81$} & \textbf{$93.69$} & \textbf{$95.06$} & \textbf{$94.57$} & \textbf{$91.57$} & \textbf{$94.62$} & \textbf{$91.75$} \\
 & Caucasian & $33.99$ & $47.77$ & $39.84$ & $76.66$ & $70.81$ & $58.57$ & $81.22$ & $73.89$ & $75.87$ & $83.50$ \\
 & Indian & $37.68$ & $53.21$ & $65.32$ & $67.52$ & $78.83$ & $83.85$ & $77.38$ & $88.51$ & $89.92$ & $83.10$ \\
 & Asian & $32.59$ & $76.39$ & $86.61$ & $83.28$ & $86.01$ & $87.75$ & $87.50$ & $84.87$ & $85.42$ & $91.52$ \\ \midrule
\multirow{4}{*}{\shortstack[l]{VGGSNN\\(TET=True)}} 
 & \textbf{African} & \textbf{$93.60$} & \textbf{$96.20$} & \textbf{$94.88$} & \textbf{$94.62$} & \textbf{$97.97$} & \textbf{$97.09$} & \textbf{$95.37$} & \textbf{$98.23$} & \textbf{$96.78$} & \textbf{$95.98$} \\
 & Caucasian & $52.08$ & $69.72$ & $79.09$ & $83.85$ & $75.57$ & $83.70$ & $91.18$ & $83.89$ & $87.86$ & $88.40$ \\
 & Indian & $77.42$ & $68.18$ & $80.41$ & $76.85$ & $76.76$ & $74.08$ & $82.13$ & $84.68$ & $80.81$ & $84.55$ \\
 & Asian & $66.12$ & $86.76$ & $84.72$ & $89.14$ & $94.49$ & $94.00$ & $90.58$ & $93.06$ & $94.99$ & $94.89$ \\ \midrule
\multirow{4}{*}{\shortstack[l]{VGGSNN\\(noTET)}} 
 & \textbf{African} & \textbf{$87.38$} & \textbf{$90.78$} & \textbf{$94.53$} & \textbf{$94.31$} & \textbf{$96.07$} & \textbf{$96.29$} & \textbf{$93.20$} & \textbf{$96.47$} & \textbf{$95.32$} & \textbf{$96.65$} \\
 & Caucasian & $46.73$ & $80.57$ & $81.22$ & $83.94$ & $69.18$ & $81.02$ & $86.27$ & $83.20$ & $85.58$ & $90.58$ \\
 & Indian & $78.21$ & $64.61$ & $73.11$ & $76.85$ & $82.04$ & $76.89$ & $87.28$ & $87.72$ & $84.60$ & $78.48$ \\
 & Asian & $74.26$ & $85.12$ & $87.30$ & $84.87$ & $93.95$ & $94.94$ & $89.73$ & $91.22$ & $94.59$ & $94.10$ \\ \bottomrule
\end{tabular}%
}
\end{table}

\begin{table}[t]
\centering
\caption{Evolution of Test Accuracy (\%) over the First 10 Epochs on the Original \textbf{RFW} Dataset.}
\label{tab:traj_rfw}
\resizebox{\textwidth}{!}{%
\begin{tabular}{@{}llcccccccccc@{}}
\toprule
\textbf{Model \& Setting} & \textbf{Demographic} & \textbf{E0} & \textbf{E1} & \textbf{E2} & \textbf{E3} & \textbf{E4} & \textbf{E5} & \textbf{E6} & \textbf{E7} & \textbf{E8} & \textbf{E9} \\ \midrule
\multirow{4}{*}{\shortstack[l]{Spike-ResNet19\\(TET=True)}} 
 & \textbf{African} & \textbf{$79.85$} & \textbf{$89.01$} & \textbf{$96.13$} & \textbf{$94.17$} & \textbf{$93.09$} & \textbf{$90.62$} & \textbf{$96.81$} & \textbf{$96.92$} & \textbf{$96.35$} & \textbf{$97.82$} \\
 & Caucasian & $47.16$ & $49.45$ & $67.86$ & $81.34$ & $76.64$ & $85.28$ & $81.05$ & $75.84$ & $79.37$ & $84.58$ \\
 & Indian & $4.50$ & $61.24$ & $67.21$ & $58.41$ & $72.95$ & $77.56$ & $70.04$ & $74.76$ & $84.21$ & $66.72$ \\
 & Asian & $66.15$ & $74.60$ & $72.42$ & $80.12$ & $81.43$ & $71.39$ & $86.11$ & $88.97$ & $85.08$ & $91.15$ \\ \midrule
\multirow{4}{*}{\shortstack[l]{Spike-ResNet19\\(noTET)}} 
 & \textbf{African} & \textbf{$78.74$} & \textbf{$91.73$} & \textbf{$96.53$} & \textbf{$91.23$} & \textbf{$92.13$} & \textbf{$86.36$} & \textbf{$94.88$} & \textbf{$95.31$} & \textbf{$97.39$} & \textbf{$95.67$} \\
 & Caucasian & $66.62$ & $38.37$ & $55.72$ & $75.15$ & $76.17$ & $83.05$ & $79.99$ & $60.02$ & $78.24$ & $87.21$ \\
 & Indian & $10.80$ & $70.65$ & $73.14$ & $74.20$ & $73.18$ & $76.65$ & $72.91$ & $84.13$ & $84.10$ & $77.94$ \\
 & Asian & $47.22$ & $77.18$ & $68.97$ & $77.26$ & $82.22$ & $78.41$ & $84.37$ & $90.28$ & $85.24$ & $89.21$ \\ \midrule
\multirow{4}{*}{\shortstack[l]{VGGSNN\\(TET=True)}} 
 & \textbf{African} & \textbf{$90.59$} & \textbf{$97.21$} & \textbf{$94.45$} & \textbf{$96.85$} & \textbf{$95.38$} & \textbf{$97.57$} & \textbf{$95.78$} & \textbf{$96.56$} & \textbf{$97.32$} & \textbf{$96.35$} \\
 & Caucasian & $71.36$ & $73.07$ & $83.05$ & $68.11$ & $87.06$ & $82.91$ & $83.38$ & $77.44$ & $87.68$ & $88.92$ \\
 & Indian & $67.09$ & $71.29$ & $76.88$ & $86.06$ & $76.28$ & $83.45$ & $87.57$ & $92.07$ & $85.80$ & $84.85$ \\
 & Asian & $65.87$ & $80.83$ & $78.17$ & $85.08$ & $90.28$ & $90.63$ & $90.60$ & $90.16$ & $91.43$ & $91.35$ \\ \midrule
\multirow{4}{*}{\shortstack[l]{VGGSNN\\(noTET)}} 
 & \textbf{African} & \textbf{$93.27$} & \textbf{$97.96$} & \textbf{$96.81$} & \textbf{$96.13$} & \textbf{$95.10$} & \textbf{$96.49$} & \textbf{$96.89$} & \textbf{$97.75$} & \textbf{$95.24$} & \textbf{$96.21$} \\
 & Caucasian & $66.51$ & $61.30$ & $81.45$ & $80.79$ & $87.57$ & $85.35$ & $81.71$ & $73.91$ & $91.87$ & $77.84$ \\
 & Indian & $68.79$ & $70.08$ & $77.67$ & $78.58$ & $67.96$ & $78.54$ & $85.27$ & $92.63$ & $77.11$ & $88.70$ \\
 & Asian & $66.55$ & $81.03$ & $73.29$ & $82.98$ & $88.21$ & $88.93$ & $90.83$ & $86.75$ & $91.31$ & $92.18$ \\ \bottomrule
\end{tabular}%
}
\end{table}


\end{document}